\documentclass{article}

\usepackage{microtype}
\usepackage{graphicx}
\usepackage{subcaption}
\usepackage{booktabs} % for professional tables

\usepackage{hyperref}

\usepackage[preprint]{icml2026}

\usepackage{amsmath}
\usepackage{amssymb}
\usepackage{mathtools}
\usepackage{amsthm}

\usepackage[capitalize,noabbrev]{cleveref}

\theoremstyle{plain}

\theoremstyle{definition}

\theoremstyle{remark}

\usepackage[textsize=tiny]{todonotes}

\usepackage{multirow}
\usepackage[table]{xcolor}

\usepackage{xspace}
\newcommand{\methodname}{VTC-R1\xspace}
\usepackage{enumitem}
\usepackage[most]{tcolorbox}

\newcommand{\up}[1]{\ensuremath{^{\textcolor{green!70!black}{\scriptscriptstyle (+#1)}}}}
\newcommand{\down}[1]{\ensuremath{^{\textcolor{red}{\scriptscriptstyle (#1)}}}}
\newcommand{\speed}[1]{\ensuremath{^{\textcolor{green!70!black}{\scriptscriptstyle (#1\times)}}}}
\definecolor{Highlight}{rgb}{0.92,0.94,1}
\begin{document}

\twocolumn[
  \icmltitle{\methodname: Vision-Text Compression for Efficient Long-Context Reasoning}

  \icmlsetsymbol{equal}{*}

  \begin{icmlauthorlist}
    \icmlauthor{Yibo Wang}{ntu}
    \icmlauthor{Yongcheng Jing}{ntu}
    \icmlauthor{Shunyu Liu}{ntu}
    \icmlauthor{Hao Guan}{ntu}
    \icmlauthor{Rong-Cheng Tu}{ntu}\\
    \icmlauthor{Chengyu Wang}{ali}
    \icmlauthor{Jun Huang}{ali}
    \icmlauthor{Dacheng Tao}{ntu}
  \end{icmlauthorlist}

  % \vspace{4pt}
  %   \centerline{\small
  %   \includegraphics[height=1em]{figure/25231.png}\;
  %   Code: \url{https://github.com/w-yibo/VTC-R1}}

  \icmlaffiliation{ntu}{Nanyang Technical University}
  \icmlaffiliation{ali}{Alibaba Cloud Computing}

  % \icmlcorrespondingauthor{Dacheng Tao}{dacheng.tao@ntu.edu.sg}

  \icmlkeywords{Machine Learning, ICML}

  \vskip 0.3in
]



\printAffiliationsAndNotice{}  % no special notice (required even if empty)

\begin{abstract}
Long-context reasoning has significantly empowered large language models (LLMs) to tackle complex tasks, yet it introduces severe efficiency bottlenecks due to the computational complexity. Existing efficient approaches often rely on complex additional training or external models for compression, which limits scalability and discards critical fine-grained information. In this paper, we propose \methodname, a new efficient reasoning paradigm that integrates vision-text compression into the reasoning process. Instead of processing lengthy textual traces, \methodname renders intermediate reasoning segments into compact images, which are iteratively fed back into vision-language models as "optical memory." We construct a training dataset based on OpenR1-Math-220K achieving 3.4× token compression and fine-tune representative VLMs--Glyph and Qwen3-VL. Extensive experiments on benchmarks such as MATH500, AIME25, AMC23 and GPQA-D demonstrate that \methodname consistently outperforms standard long-context reasoning. Furthermore, our approach significantly improves inference efficiency, achieving 2.7× speedup in end-to-end latency, highlighting its potential as a scalable solution for reasoning-intensive applications. Our code is available at \url{https://github.com/w-yibo/VTC-R1}.
\end{abstract}

\section{Introduction}
\label{sec:introduction}

\begin{figure}[t]
    \centering
    \includegraphics[width=\linewidth]{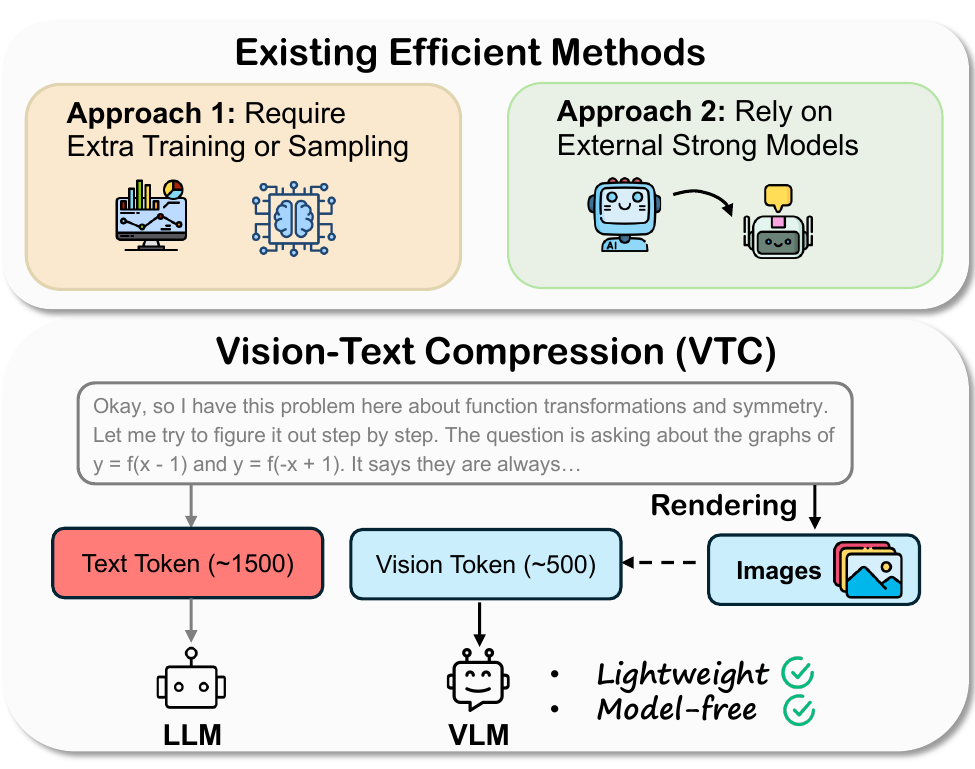}
    \caption{\textbf{Comparison between existing efficient reasoning approaches and vision-text compression (VTC).} Existing methods either require additional training or sampling procedures, or rely on external strong models. In contrast, VTC leverages lightweight rendering to transform long textual reasoning traces into compact visual representations, enabling VLMs to encode information with significantly fewer vision tokens (3-4× compression). This approach is both lightweight and model-free.}
    \label{fig:comparison}
    \vspace{-4mm}
\end{figure}

Reasoning capability~\cite{sys1tosys2,lightman2023lets,yao2023tree,huang2023towards, yao2025survey} has emerged as a powerful technique of large language models (LLMs), enabling them to tackle complex tasks such as mathematical problem solving~\cite{hendrycksmath2021, adar1,hu2025beyond} and code generation~\cite{llmcode, codesuvery}. Recent advancements, exemplified by OpenAI o1~\cite{o12024} and DeepSeek-R1~\cite{2025deepseekr1}, leverage reinforcement learning to further scale this capability to long-context reasoning, substantially improving performance on challenging real-world tasks~\cite{wang2026deepresearcheval}. Despite recent progress, long-context reasoning introduces severe efficiency bottlenecks. The computational complexity of the transformer architecture~\cite{zaheer2020big,beltagy2020longformer,kitaev2020reformer} grows quadratically with sequence length, causing both computation and memory costs to increase rapidly as the context expands. This leads to degraded inference speed, reduced training efficiency, and limited scalability, which significantly hinders real-world deployment.

To mitigate these issues, several efficient approaches are proposed~\cite{overthinking,munkhbat2025selftrainingelicitsconcisereasoning,lee2025well,liu2024expediting}. Existing methods can be broadly categorized into two groups.
i) Extra training or sampling stages beyond standard training. For example, CoT-Valve~\cite{ma2025cotvalvelengthcompressiblechainofthoughttuning} adopts a multi-stage training procedure to obtain models specialized for different reasoning lengths and O1-Pruner~\cite{luo2025o1prunerlengthharmonizingfinetuningo1like} applies offline reinforcement learning with multiple sampled trajectories (16 responses per problem). These approaches increase training  and inference cost.
ii) External strong models to guide reasoning compression. TokenSkip~\cite{xia2025tokenskipcontrollablechainofthoughtcompression} requires an additional model to estimate token importance, while R1-Compress~\cite{wang2025r1} and InftyThink~\cite{yan2025inftythink} depend on powerful external summarization models (e.g., Llama-3.3-70B-Instruct) to condense long reasoning traces. 
Although both categories of methods are effective, they often restrict exploration space and discard fine-grained information that is critical for reasoning.

\begin{quote}
 \vspace{-0.2cm}
 \centering
 \textit{Without additional training or external models, how can we achieve efficient reasoning while preserving fine-grained information?} 
  \vspace{-0.2cm}
 \end{quote}

Motivated by this, a promising yet underexplored direction is vision-text compression (VTC)~\cite{wei2025deepseek,glyph,xing2025vision,zhao2025vtcbench,xing2025see}. 
Rather than reducing fine-grained information, VTC adopts an alternative representation by transforming textual content into visual forms via lightweight rendering, enabling vision-language models (VLMs) to encode rich semantic information using substantially fewer vision tokens. 
This design is lightweight and model-free, as shown in Figure~\ref{fig:comparison}, introducing no additional training stages or reliance on external compression models.
Prior works such as DeepSeek-OCR~\cite{wei2025deepseek} and Glyph~\cite{glyph} focus on text reconstruction or long-context understanding, showing that long text sequences can be represented with 3×–10× token compression while maintaining high decoding precision. 
However, whether such high-density visual representations can preserve and support multi-step reasoning processes remains unclear.
Notably, mathematical reasoning, with its symbolic structure and step-wise derivations, is naturally amenable to visual rendering, making it a suitable and principled testbed for studying reasoning-oriented vision-text compression.

\begin{figure*}[t]
    \centering
    \includegraphics[width=0.9\linewidth]{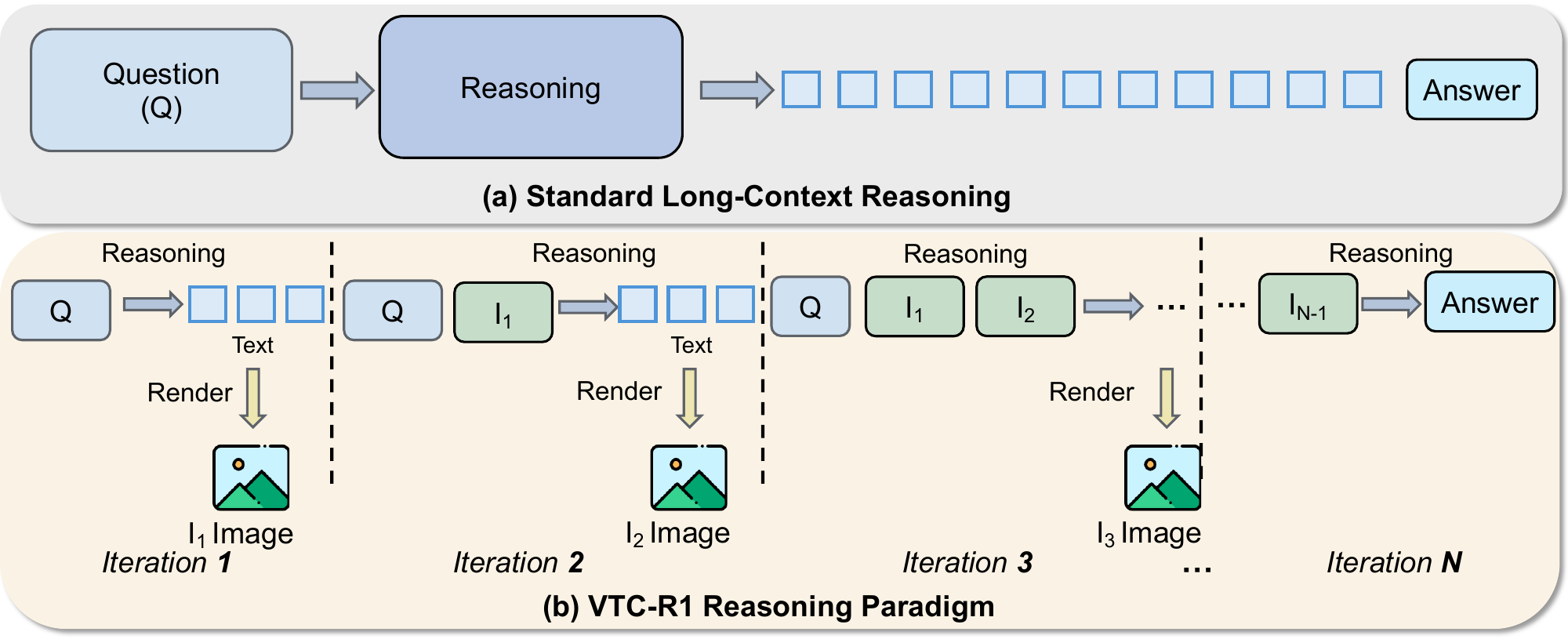}
    \caption{Comparison between standard long-context reasoning and the proposed VTC-R1 reasoning paradigm. \textbf{(a)} Standard long-context reasoning processes the entire reasoning trace as a single long sequence, leading to increasing computational and memory costs as the context grows. \textbf{(b)} VTC-R1 reformulates long-context reasoning as an iterative process. At each iteration, the current reasoning segment is generated and the preceding segments are rendered into compact images, which are fed back to the model together with the original question. These rendered images function as a form of optical memory, enabling efficient multi-step reasoning with reduced token usage.
}
    \label{fig:method}
    \vspace{-4mm}
\end{figure*}

To bridge this gap, we propose \methodname, a new efficient reasoning paradigm that iteratively integrates vision–text compression into long-context reasoning. \methodname treats the reasoning process as multiple processes, where \textit{the preceding process are regarded as long-context and rendered into compact images}, and performs iterative reasoning~\cite{yan2025inftythink} with VLMs. As illustrated in Figure \ref{fig:method}, the reasoning process is decomposed into a sequence of reasoning steps. Upon the completion of each step, it is rendered into an image. To proceed to the next step, the accumulated images of previous steps are fed back into the model alongside the question, functioning as a form of optical memory that compactly encodes previous reasoning using  vision tokens.

We construct a training dataset based on OpenR1-Math-220K~\cite{openr1}, a large-scale long-context reasoning corpus generated by DeepSeek-R1~\cite{2025deepseekr1}. We segment each long reasoning trace into shorter reasoning segments and render the preceding segments into images, forming paired image--text reasoning data with up to $3.4\times$ token compression as shown in Table~\ref{tab:stat_render}. We then fine-tune representative VTC-VLM (i.e., Glyph~\cite{glyph}) and the state-of-the-art VLM (i.e., Qwen3-VL~\cite{Qwen3-VL}), under this iterative reasoning framework. Extensive experiments on diverse mathematical reasoning benchmarks, GSM8K~\cite{cobbe2021gsm8k}, MATH500~\cite{lightman2023lets}, AIME25~\cite{aime25}, AMC23~\cite{amc23} and GPQA-Diamond~\cite{rein2024gpqa}, demonstrate that \methodname consistently outperforms standard long-context reasoning. Moreover, \methodname significantly improves inference efficiency, achieving up to 2.7× speedup in end-to-end reasoning latency, highlighting its practical advantages for scalable long-context reasoning. The main contributions of this paper:
\begin{itemize}[leftmargin=*, itemsep=0pt, parsep=0pt, topsep=0pt]
    \item We introduce \methodname, a new efficient reasoning paradigm that reformulates reasoning as an iterative process and integrates vision-text compression to replace long text with compact vision tokens, without requiring additional training stages or external strong models.

    \item We construct a training dataset by segmenting reasoning traces and rendering preceding steps into images, producing paired data with up to $3.4\times$ token compression. 
    % \item We construct a training dataset by segmenting reasoning traces and rendering preceding steps into images, producing paired image--text data with up to $3.4\times$ token compression. This process expands the original dataset into 106K reasoning instances with 105K rendered images.

    \item Extensive evaluation on major mathematical and out-of-distribution benchmarks shows that \methodname consistently outperforms standard long-context reasoning and achieves up to a 2.7x speedup in end-to-end inference latency.
\end{itemize}

\section{Related Work}

\textbf{Reasoning in Large Language Models.}
Reasoning capabilities~\cite{sys1tosys2,lightman2023lets,huang2023towards,yao2025survey} constitute a cornerstone of modern LLMs, enabling proficiency in rigorous domains like mathematics~\cite{hendrycksmath2021,adar1,hu2025beyond} and code generation~\cite{llmcode,codesuvery}. While early strategies relied on structured prompting~\cite{yao2023tree,yao2024mulberryempoweringmllmo1like}, recent advancements leverage reinforcement learning to scale test-time compute. Models such as OpenAI o1~\cite{o12024}, DeepSeek-R1~\cite{2025deepseekr1}, and Kimi~\cite{team2025kimi} generate extended chains of thought, achieving significant improvements on challenging real-world benchmarks.

\textbf{Efficient Reasoning.}
Long-context reasoning strategies exacerbate the computational bottlenecks inherent in the quadratic complexity of Transformer architectures~\cite{zaheer2020big,beltagy2020longformer,kitaev2020reformer,wang2020linformer}. Recent research has investigated various efficiency mechanisms~\cite{liu2025efficient,overthinking,munkhbat2025selftrainingelicitsconcisereasoning,lee2025well,liu2024expediting,yang2025speculativethinkingenhancingsmallmodel,zhang2025lightthinkerthinkingstepbystepcompression,hao2024traininglargelanguagemodels,yang2025dynamicearlyexitreasoning,pan2025learningadaptiveparallelreasoning,ma2025reasoningmodelseffectivethinking,qiao2025conciseconfidenceguidedcompressionstepbystep,zhuang2025acceleratingchainofthoughtreasoninggoalgradient,yang2025thinkneedselfadaptivechainofthought,hou2025thinkprunepruninglongchainofthought,ning2025thoughtsgeneratedequalefficient,li2025tldrlongreweightingefficient,gong2025efficientreasoningchainunconscious}, though existing methods often incur significant trade-offs. One category of approaches~\cite{team2025kimi}, exemplified by CoT-Valve~\cite{ma2025cotvalvelengthcompressiblechainofthoughttuning} and O1-Pruner~\cite{luo2025o1prunerlengthharmonizingfinetuningo1like}, relies on complex multi-stage training procedures or extensive offline sampling, which substantially increases pre-deployment overhead. A second category leverages external strong models~\cite{kang2024c3otgeneratingshorterchainofthought} to guide reasoning compression, as in TokenSkip~\cite{xia2025tokenskipcontrollablechainofthoughtcompression}, R1-Compress~\cite{wang2025r1}, and InftyThink~\cite{yan2025inftythink}, making the compression quality dependent on the capabilities of these auxiliary models. Although effective in reducing token counts, these approaches often constrain the exploration space and risk discarding fine-grained information that is critical for correct logical deduction.

\textbf{Vision-Text Compression.}
Vision-text compression (VTC)~\cite{wei2025deepseek,glyph,xing2025vision,xing2025see,wang2024visincontext} has emerged as a promising approach for reducing the cost of processing long textual sequences by transforming text into compact visual representations. DeepSeek-OCR~\cite{wei2025deepseek} demonstrates that long texts can be compressed into visual tokens, achieving a 3$\times$--10$\times$ reduction in token count while maintaining high decoding fidelity. Glyph~\cite{glyph} utilizes continuous pre-training and RL for VTC to enhance long-context understanding capabilities. VTCBench~\cite{zhao2025vtcbench} proposes a benchmark to evaluate the spectrum of capabilities in VTC. While prior work focuses on text understanding and reconstruction, and it remains unclear whether such high-density visual representations can faithfully preserve and support complex reasoning processes, particularly for mathematically intensive and multi-step reasoning tasks. 

Some concurrent works, AgentOCR~\cite{feng2026agentocr} utilizes VTC to compress the agent's history derived from tool invocations into a compact rendered image. RoT~\cite{wang2026render} focuses on utilizing rendered visual tokens as latent tokens for latent reasoning, but it does not explicitly address long-context reasoning and lacks systematic evaluation on challenging benchmarks.

\section{Preliminaries}

\subsection{Problem Setup}
We consider a reasoning task defined by an input question $Q$. Given a vision language model $M$, the goal is to produce a final answer $A$. During answer generation, a long sequence of intermediate reasoning steps is also produced, which forms a \textbf{long-context reasoning}.

\subsection{Vision-Text Compression}
\label{subsec:vtc}
Vision-text compression is defined as a procedure where a given text is rendered into an image, enabling a VLM to encode the content using fewer vision tokens. The pipeline used in our work is summarized as follows.

Given an input text sequence $T$, the text is rendered into images through a pipeline before model input. The rendering pipeline is parameterized by a configuration vector~\cite{glyph},
\begin{equation}
\begin{aligned}
\theta = \big(&
\text{dpi}, \text{page\_size}, \text{font\_family}, \text{font\_size},\\
&\text{line\_height}, \text{alignment},
\text{indent}, \text{spacing},\\
&\text{h\_scale}, \text{colors}, \text{borders}, \ldots\big),
\end{aligned}
\end{equation}
which controls the typography, layout, and visual style of rendered pages. The details of rendering configuration are provided in Appendix~\ref{apd:rendering_config}. Through the rendering process, multiple PNG images $I$ are produced. This process is defined as
$I = R_\theta(T)$,
where $R_\theta(\cdot)$ denotes the rendering operator. The images $I$ are processed by the image processor and vision encoder of model $M$.
For simplicity, let $M_{\text{vision}}$ denote the vision tokenizer. Given the images $I$, we obtain a sequence of vision tokens $V = M_{\text{vision}}(I)$, where $V = \{v_1, \ldots, v_{L_v}\}$ and $L_v$ represents the sequence length.

The original text $T$ is processed by the text tokenizer $M_{\text{txt}}$ to produce a text token sequence $T = M_{\text{txt}}(T)$, where $T = \{t_1, \ldots, t_{L_t}\}$ and $L_t$ denotes the number of text tokens.

Thus, the vision-text compression ratio is defined as:
\begin{equation}
    \rho = \frac{L_t}{L_v},
    \label{eq:compression_ratio}
\end{equation}
In practice, $\rho > 1$, a larger $\rho$ indicates a higher compression efficiency, implying that fewer tokens are required to encode the same content under the vision tokenization scheme.

\section{Methodology}

\subsection{Standard Long-Context Reasoning.}

Standard long-context reasoning, as adopted by OpenAI o1~\cite{o12024} and DeepSeek-R1~\cite{2025deepseekr1}, typically produces a long sequence of intermediate reasoning steps. Such behavior incurs substantial computational and memory cost. This reasoning procedure is formulated as a long-context reasoning process, denoted as $LR$, where the input question is $Q$. The standard long-context reasoning can be represented as\[
\begin{aligned}
\langle \texttt{$S_{s}$} \rangle \,|\, \texttt{U} \,|\, Q \,|\, 
\texttt{A} \,|\, 
\langle \texttt{think} \rangle \, LR \, \langle / \texttt{think} \rangle \, A ,
\end{aligned}
\]

where $\langle \texttt{$S_{s}$} \rangle$ denotes the standard system prompt, such as ``You are a helpful assistant.''  
The tokens $\,|\, \texttt{U} \,|\,$ and $\,|\, \texttt{A} \,|\,$ indicate the start of user input and model response, respectively.  
The special tokens $\langle \texttt{think} \rangle$ and $\langle / \texttt{think} \rangle$ mark the beginning and end of the reasoning process.

In practice, $LR$ may reach 16k tokens or more. During reasoning, the preceding steps could be \textbf{regarded as context} and vision-text compression can therefore be introduced to encode these preceding steps into \textbf{a smaller number of effective vision tokens}, thereby mitigating the substantial cost of long-context reasoning.

% Our method instead represents previous reasoning steps using compact visual tokens and performs iterative reasoning over mixed vision-text inputs.

\begin{algorithm}[t]
\caption{\methodname Reasoning Paradigm}
\label{algo:vtc_r1}
\begin{algorithmic}
\STATE {\bfseries Input:} question $Q$; vision language model $M$; system prompt $\langle \texttt{S}_{v} \rangle$; rendering operator $R_\theta$; maximum iteration $T$
\STATE {\bfseries Initialize:} rendered image set $\mathcal{I} \leftarrow \emptyset$

\FOR{$i = 1$ to $T$}
    \STATE \textcolor{gray}{Generate Vision-Language Model Output:}
    \[
    O_i \leftarrow M(\langle \texttt{S}_{v} \rangle, Q, \mathcal{I})
    \]

    \IF{$O_i$ produces the final answer $A$}
        \STATE \textbf{return} $A$
    \ENDIF

    \STATE \textcolor{gray}{Update Image Set via Rendering:}
    \STATE Extract reasoning progress $LR_i$ from $O_i$
    \STATE Render reasoning into images: $I_i \leftarrow R_\theta(LR_i)$
    \STATE Update: $\mathcal{I} \leftarrow \mathcal{I} \cup \{I_i\}$
\ENDFOR

\IF{no final answer $A$ }
    \STATE \textcolor{gray}{Extract Answer when Reaching Iteration Limit:}
    \STATE Extract final answer $A$ from $O_T$
\ENDIF
\STATE {\bfseries Output:} final answer $A$
\end{algorithmic}
\end{algorithm}

\subsection{\methodname Reasoning}
\label{subsec:vtc-r1}

Instead of generating a full textual reasoning trace, VTC-R1 first formulates long-context reasoning as an iterative process to get the answer. 
A long-context reasoning process, denoted as $LP$, is decomposed into a sequence of reasoning segments $\{LP_1, \ldots, LP_n\}$. 

\textbf{Iterative Reasoning.}
% \yibo{need be check, and can be longer after polishing}
Concretely, iterative reasoning generates the reasoning process sequentially. At iteration $i$, the model conditions on the question and the previous segments:
\begin{equation}
LP_i \sim p_\theta(\cdot \mid Q, LP_{<i}), \qquad LP_{<i}\triangleq (LP_1,\ldots,LP_{i-1}),
\end{equation}
and the complete trace is obtained by concatenation $LP = (LP_1,\ldots,LP_n)$.

We next show that \textit{this iterative formulation is equivalent to standard one-pass long-context generation under an autoregressive model}. By the chain rule, the joint distribution of the full trace factorizes as
\begin{equation}
p_\theta(LP\mid Q)=\prod_{i=1}^{n} p_\theta(LP_i \mid Q, LP_{<i}),
\end{equation}
which is exactly the distribution induced by sampling $LP_1,\ldots,LP_n$ sequentially with the same conditionals. Consequently, for any answer extraction function $A=M(LP)$, both one-pass and iterative generation yield the same answer distribution:
\[
A = M(LP), \quad LP \sim p_\theta(\cdot\mid Q).
\]

\textbf{\methodname Reasoning Paradigm.}
The first reasoning process is expressed as follows, where $n>1$ is assumed:
\[
\begin{aligned}
\langle \texttt{S}_{v} \rangle \,|\, \texttt{U} \,|\, Q \,|\, 
\texttt{A} \,|\, 
\langle \texttt{think} \rangle \, LR_1 \, \langle / \texttt{think} \rangle ,
\end{aligned}
\]
where $\langle \texttt{S}_{v} \rangle$ denotes the VTC-R1 system prompt.
\begin{tcolorbox}[
    colback=gray!20,
    colframe=gray!60,
    boxrule=0.5pt,
    arc=3pt,
    left=6pt,
    right=6pt,
    top=6pt,
    bottom=6pt
]
\textbf{VTC-R1 System Prompt }$\langle \texttt{S}_{v} \rangle$ \\[2pt]
These images record your previous reasoning process.
Based on this reasoning, continue and complete the final answer.
Do not restart the reasoning.

If no images are provided, start the reasoning from scratch.
\end{tcolorbox}

As described in Sec~\ref{subsec:vtc}, the first reasoning process $LR_1$ is rendered into multiple images, $I_1 = R_\theta(LR_1)$.

% When the $i$-th reasoinng process begins, where $1<i<n$, a set of rendered images 
% $\{I_1, \ldots, I_{i-1}\}$ is available.
When the $i$-th reasoning process begins, $i-1$ reasoning processes have been completed. 
At the end of each process, the generated reasoning process $LR_j$ is rendered into multiple images $I_j$ and stored. 
As a result, a set of rendered images $\{I_1, \ldots, I_{i-1}\}$ is available.
The reasoning process at the $i$-th iteration is then expressed as
\[
\begin{aligned}
\langle \texttt{S}_{v} \rangle \,|\, \texttt{U} \,|\, Q,\, I_1, \ldots, I_{i-1} \,|\, 
\texttt{A} \,|\, 
\langle \texttt{think} \rangle \, LR_i \, \langle / \texttt{think} \rangle .
\end{aligned}
\]

At the final reasoning iteration $n$, the model produces the last reasoning segment and outputs the final answer $A$. The complete generation at this stage is expressed as
\[ \begin{aligned} \langle \texttt{S}_{v} \rangle \,|\, \texttt{U} \,|\, Q,\, I_1, \ldots, I_{n-1} \,|\, \texttt{A} \,|\, \langle \texttt{think} \rangle \, LR_n \, \langle / \texttt{think} \rangle A. \end{aligned} \]

During inference, \methodname iterates continuously until the final answer $A$ is produced. 
As shown in Table~\ref{tab:math_benchmark_main}, the method exhibits adaptive reasoning behavior, where the number of reasoning iterations is selected dynamically according to the problem difficulty. 
To prevent unbounded generation, a maximum iteration limit, denoted as $T$, is imposed.

\methodname performs iterative reasoning by generating multiple reasoning segments in Algorithm~\ref{algo:vtc_r1}. 
At each iteration, the previously generated reasoning segments $LR_1, \ldots, LR_{i-1}$ are rendered into images $I_1, \ldots, I_{i-1}$. 
Therefore, these images provide a compact and efficient representation of textual reasoning through vision tokens, functioning analogously to an optical memory. 
Under our rendering configuration, the resulting \textbf{compression ratio $\rho$ is approximately $3$--$4$} as shown in Table~\ref{tab:stat_render}, which could mitigate the computational and memory cost incurred by token growth in standard long-context reasoning.

Moreover, \methodname requires a lightweight rendering mechanism. 
No additional training, extra sampling stages, or external models are introduced.

\textbf{Batch Inference}.
To facilitate batch inference in frameworks like vLLM~\cite{kwon2023efficient}, we adapt Algorithm~\ref{algo:vtc_r1} by introducing independent \textit{request states} and a \textit{dynamic active set} mechanism. This approach enables efficient parallel generation by selectively constructing batch inputs and updating multimodal contexts only for active samples during each iteration. The detailed Algorithm~\ref{algo:vtc_r1_batch} is provided.

\begin{figure}[t]
    \centering
    \includegraphics[width=\linewidth]{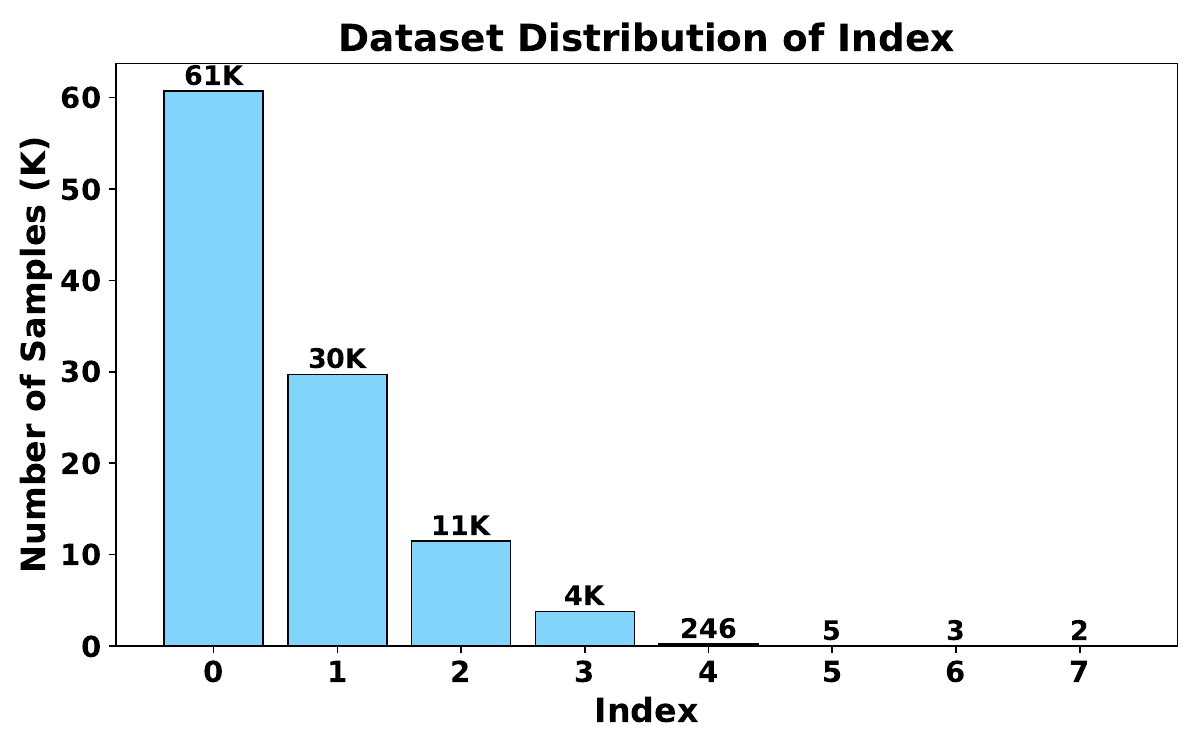}
    \caption{\textbf{Distribution of data index.}
    The index indicates the order of a reasoning segment for a given problem, where index $0$ corresponds to the first segment.
    Most samples terminate at early steps, while a small fraction requires more than four iterations.}

    \label{fig:index_distribution}
    \vspace{-2mm}
\end{figure}

\begin{table}[t]
\centering
\caption{\textbf{Statistics of rendered prior reasoning segments.}
We report the number of reasoning segments rendered as images, the total number of text and vision tokens, and the compression ratio.}
\label{tab:vtc_statistics}
\begin{tabular}{l c}
\toprule
\textbf{Metric}& \textbf{Value} \\
\midrule
Rendered reasoning steps (elements)        & 45K \\
Rendered images                            & 105K \\
Total text tokens& 181M \\
Total vision tokens                        & 54M \\
Compression ratio (text / vision)          & \textbf{3.4$\times$} \\
\bottomrule
\label{tab:stat_render}
\end{tabular}
\vspace{-6mm}
\end{table}

\begin{table*}[t]
\centering
% \small
\caption{\textbf{Performance comparison across mathematical benchmarks.}
Accuracy (ACC) is higher-is-better~($\uparrow$), latency (LAT) is lower-is-better~($\downarrow$).
\textbf{Bold} indicates the best performance.
Superscript numbers denote accuracy improvements and latency speedups relative to standard long-context reasoning.}
\label{tab:math_benchmark_main}
\resizebox{1.0\linewidth}{!}{
\begin{tabular}{l|ccc|ccc|ccc|ccc}
\toprule
\multirow{2}{*}{\textbf{Model}} 
& \multicolumn{3}{c}{\textbf{GSM8K}}
& \multicolumn{3}{c}{\textbf{MATH500}}
& \multicolumn{3}{c}{\textbf{AIME25} (Avg@16)}
& \multicolumn{3}{c}{\textbf{AMC23} (Avg@16)}\\
\cmidrule(lr){2-4} \cmidrule(lr){5-7} \cmidrule(lr){8-10} \cmidrule(lr){11-13} 
& ACC $\uparrow$ & TOK & LAT$\downarrow$
& ACC $\uparrow$ & TOK & LAT$\downarrow$
& ACC $\uparrow$ & TOK & LAT$\downarrow$
& ACC $\uparrow$ & TOK & LAT$\downarrow$\\

\midrule
\multicolumn{13}{l}{\textit{Qwen3-VL-8B}}\\
SFT 
& 88.1 & 1.79 & 3.04 
& 85.4 & 4.17 & 5.36 
& \textbf{32.71} & 17.46 & 29.85 
& 75.00 & 8.20 & 11.08 \\

\rowcolor{Highlight}
\textbf{\methodname} 
& \textbf{94.7}\up{6.6} & 1.09 & \textbf{0.46}\speed{6.6} 
& \textbf{90.0}\up{4.6} & 3.39 & \textbf{2.49}\speed{2.2} 
& 30.00\down{-2.71} & 14.32 & \textbf{12.02}\speed{2.5} 
& \textbf{77.97}\up{2.97} & 8.18 & \textbf{6.45}\speed{1.7}  \\

\midrule
\multicolumn{13}{l}{\textit{Glyph}}\\
Base SFT 
& 86.1 & 2.35 & 1.38 
& 79.6 & 5.51 & 2.77 
& 24.17 & 19.94 & 14.48 
& 61.56 & 12.67 & 8.55 \\

SFT 
& 87.1 & 1.87 & 0.93 
& 80.4 & 5.71 & 3.05 
& 25.62 & 17.47 & 11.52 
& 60.94 & 11.65 & 6.85\\

TokenSkip 
& 86.4 & 2.25 & 1.32 
& 80.6 & 6.11 & 3.05 
& 23.75 & 17.82 & 11.85 
& 59.53 & 12.81 & 8.41\\

\rowcolor{Highlight}
\textbf{\methodname} 
& \textbf{93.6}\up{6.5} & 1.09 & \textbf{0.34}\speed{2.7} 
& \textbf{86.0}\up{5.6} & 4.12 & \textbf{2.19}\speed{1.4}
& \textbf{26.25}\up{0.63} & 12.95 & \textbf{6.81}\speed{1.7}
& \textbf{64.38}\up{3.44} & 8.81 & \textbf{4.30}\speed{1.6}\\

\bottomrule
\end{tabular}
}
\vspace{-4mm}
\end{table*}

\subsection{Training Data Construction}

To train \methodname, a supervised fine-tuning dataset is constructed to enable VLMs to learn the \methodname reasoning paradigm. 
The dataset is organized as an image--text paired corpus.
We adopt OpenR1-Math-Inf~\cite{yan2025inftythink}, which is a subset of OpenR1-Math-220K~\cite{openr1}.
OpenR1-Math-220K is generated by the DeepSeek-R1~\cite{2025deepseekr1} model, where solutions are produced for large-scale mathematical problems.
OpenR1-Math-Inf contains 61K question--answer pairs, and each solution is partitioned into multiple reasoning segments
$\{ LR_1, LR_2, \ldots, LR_n \}$ according to predefined thresholds.

Based on Sec~\ref{subsec:vtc-r1}, training data are constructed according to the index of the reasoning process, where different rules are applied at different iterations. 
Rendered images are included as inputs.
The instance at iteration $i$ is defined as
\begin{equation}
{Data}_i =
\begin{cases}
\bigl(
\langle \texttt{S}_v \rangle,
Q,
\varnothing,
LR_1
\bigr),
& i = 1, \\[6pt]

\bigl(
\langle \texttt{S}_v \rangle,
Q,
\{ I_j \}_{j<i},
LR_i
\bigr),
& 1 < i < n, \\[6pt]

\bigl(
\langle \texttt{S}_v \rangle,
Q,
\{ I_j \}_{j<i},
LR_n,
A
\bigr),
& i = n .
\end{cases}
\label{eq:data_construction}
\end{equation}
106K instances are constructed based on Eq.~\ref{eq:data_construction}, which requires approximately 105K rendered images in PNG format.
Figure~\ref{fig:index_distribution} presents the segment index distribution in the constructed training data.
Table~\ref{tab:stat_render} reports the token statistics after applying vision-text compression.
The original reasoning traces contain 181M text tokens, which are reduced to 54M vision tokens after rendering, achieving a compression ratio of up to \textbf{3.4$\times$}.
This dataset is subsequently used for supervised fine-tuning.
It is noted that the number of images associated with each instance is adaptive. 
Therefore, the training procedure requires VLM architectures that support inputs with a variable number and resolution of images, such as Qwen3-VL~\cite{Qwen3-VL}, GLM-4.1V~\cite{vteam2025glm45vglm41vthinkingversatilemultimodal}, and Glyph~\cite{glyph}.

% \subsection{Inference}

\section{Experiments}

\begin{figure*}[t]
\centering
\includegraphics[width=\textwidth]{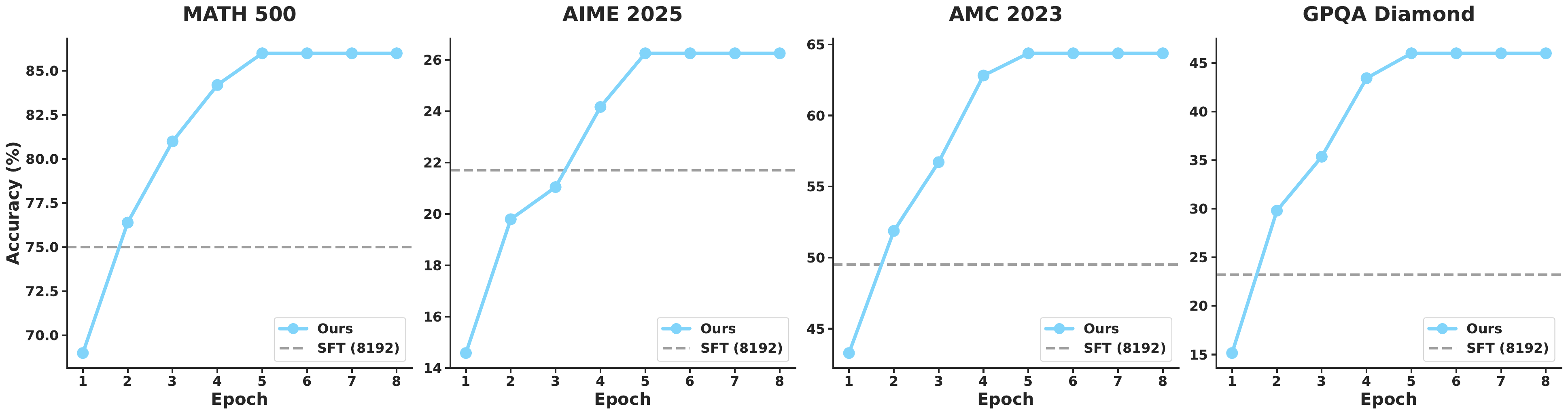}
% \vspace{1mm}
\caption{Accuracy of the proposed method across benchmarks under different maximum iteration epochs. The epoch index denotes the maximum number of allowed reasoning iterations, and predictions that terminate earlier are also included in evaluation.
The dashed line indicates the single-round baseline (standard long-context reasoning for 8192 maximum tokens).
}
\label{fig:training_dynamics}
\vspace{-4mm}
\end{figure*}

\subsection{Experiment Settings}

\textbf{Dataset.} 
For training, we use the OpenR1-Math-Inf~\cite{yan2025inftythink}, where each solution is segmented into multiple reasoning segments with varying lengths (2K, 4K, and 6K tokens). it is a subset of OpenR1-Math-220K~\cite{openr1} dataset. Unless otherwise specified, 4K is used as the default segmentation setting.
For evaluation, we leverage four widely used mathematical reasoning benchmarks. GSM8K~\cite{cobbe2021gsm8k}, MATH500~\cite{lightman2023lets}, AIME25~\cite{aime25}  and AMC23~\cite{amc23}. And GPQA-Diamond (GPQA-D)~\cite{rein2024gpqa}, a science-domain benchmark that serves as an out-of-distribution evaluation. See the Appendix~\ref{apd:subsec_bench} for more details of benchmarks.

\textbf{Baseline.}
For baselines, our proposed method \methodname is compared with standard long-context reasoning (SFT). 
In the SFT setting, standard question–answer pairs with full long-form reasoning traces are used as the supervised fine-tuning dataset. 
We then perform \methodname and SFT on two representative VLM architectures respectively for comparison. 
i) Glyph~\cite{glyph}, which serves as a VTC-capable VLM. ii) Qwen3-VL-8B~\cite{Qwen3-VL}, which represents a mainstream vision–language model.
In addition, standard SFT does not require optical character recognition capability. 
Therefore, the base model preceding Glyph, GLM-4.1V-9B-Base~\cite{vteam2025glm45vglm41vthinkingversatilemultimodal} (Base SFT), is also included as a baseline. The efficient reasoning method TokenSkip~\cite{xia2025tokenskipcontrollablechainofthoughtcompression} is included as an additional baseline for comparison.

\textbf{Metric.}
We employ the following three metrics to evaluate the model’s performance. 

\begin{itemize}[leftmargin=*, itemsep=0pt, parsep=0pt, topsep=0pt]
    \item \textbf{Accuracy (ACC)}: 
    For GSM8K, MATH500, and GPQA-Diamond, we report pass@1 accuracy. 
    For AIME25 and AMC23, due to their limited dataset sizes, we generate 16 responses per problem and report avg@16 accuracy.
    \item \textbf{Token (TOK)}: 
    The average number of tokens in the generated responses.
    \item \textbf{Latency (LAT)}: 
    We measure the average inference latency per generation. 
    Given a dataset with $m$ problems, where each problem is generated $n$ times (e.g., $n=16$ for AIME25 and AMC23), let $t_1$ and $t_2$ denote the wall-clock timestamps at the start and end of the entire inference process, respectively. 
    The latency is computed as:
    \[
        LAT = \frac{t_2 - t_1}{m \times n}.
    \]
\end{itemize}

\textbf{Implementation Details.} 
For both SFT and \methodname, the processed training datasets contain 106K instances and require approximately 105K images. 
Both methods are trained with a learning rate of $1\times10^{-5}$ for one epoch using the \texttt{LlamaFactory} library~\cite{zheng2024llamafactory}. 
For evaluation, a temperature of $0.6$ and a top-$p$ value of $0.95$ are adopted under the vLLM framework~\cite{kwon2023efficient}.

\subsection{Main Results}

\textbf{Performance Gains.}
As shown in Table~\ref{tab:math_benchmark_main}, \methodname consistently outperforms Base SFT, SFT and TokenSkip baselines on the Glyph across all four benchmarks. 
Notably, substantial improvements are observed on the more challenging benchmarks, with the gains of 5.6\% on MATH500 and 3.4\% on AMC23.
On the Qwen3-VL architecture, \methodname also demonstrates consistent improvements or achieves competitive accuracy compared to standard long-context reasoning.

Furthermore, as reported in Table~\ref{tab:gpqa_results}, similar trends are observed on the out-of-distribution benchmark.
Specifically, \methodname yields accuracy improvements of 7.6\% and 11.1\%, indicating that the proposed approach generalizes effectively beyond in-distribution mathematical benchmarks.

\textbf{Efficient Inference.}
\methodname achieves efficient inference latency across model architectures.
On the Glyph architecture, a speedup of at least $1.4\times$ is observed across all benchmarks, with larger gains of $1.7\times$ and $1.6\times$ on the more challenging benchmarks.
On the Qwen3-VL architecture, the inference speedup reaches up to $6.6\times$.

Although the proposed method is not explicitly designed as an adaptive reasoning framework, adaptivity naturally emerges from the data construction process, where different problems are associated with different numbers of reasoning iterations.
As a result, benchmarks of varying difficulty exhibit different effective token lengths (TOK).
For instance, GSM8K requires fewer tokens, while AIME25 involves longer token sequences and more iteration epochs.

The latency speedup consistently exceeds the reduction in token count.
For example, on the Glyph for AMC23, the token count is reduced by approximately $1.3\times$, whereas the latency improvement reaches $1.6\times$.
This discrepancy indicates that the introduction of vision-text compression provides additional efficiency gains beyond token reduction.

\begin{table}[t]
\centering
\small
\caption{Performance on Out-of-distribution Benchmark (GPQA-Diamond).
\textbf{Bold} indicates the best performance.}
\label{tab:gpqa_results}
\setlength{\tabcolsep}{10pt}
\begin{tabular}{l|ccc}
\toprule
\multirow{2}{*}{\textbf{Model}} 
& \multicolumn{3}{c}{\textbf{GPQA-Diamond}}\\
\cmidrule(lr){2-4}
& ACC $\uparrow$ & TOK & LAT$\downarrow$\\

\midrule
\multicolumn{4}{l}{\textit{Qwen3-VL-8B}}\\
SFT 
& 37.4 & 14.78 & 26.88 \\
\rowcolor{Highlight}
\textbf{\methodname} 
& \textbf{48.5}\up{11.1} & 9.77 & \textbf{9.57}\speed{2.8} \\

\midrule
\multicolumn{4}{l}{\textit{Glyph}}\\
Base SFT 
& 26.3 & 19.74 & 14.43 \\
SFT 
& 38.4 & 13.91 & 8.35 \\

TokenSkip
& 35.9 & 15.45 & 9.93\\

\rowcolor{Highlight}
\textbf{\methodname} 
& \textbf{46.0}\up{7.6} & 10.73 & \textbf{6.96}\speed{1.2}\\

\bottomrule
\end{tabular}

\vspace{-4mm}
\end{table}

% \subsection{Iteration Epochs}

\textbf{Iteration Epochs.}
Figure~\ref{fig:training_dynamics} illustrates the accuracy of the proposed method across four benchmarks over different epoch settings. 
Here, the epoch index denotes the maximum number of allowed reasoning iterations. 
Predictions that terminate before reaching the maximum epoch are also included in the evaluation, which results in a non-decreasing accuracy trend as the epoch limit increases. As shown in the figure, the accuracy consistently improves as the maximum epoch increases, demonstrating the effectiveness of multi-iteration reasoning. 
Across most benchmarks, the rate of accuracy improvement gradually diminishes, and performance begins to converge from approximately the fifth epoch onward. 
This observation indicates that the proposed method benefits from additional reasoning iterations while exhibiting stable convergence behavior.

\textbf{Overcoming Training Context Limitations.}
The gray dashed line in Figure~\ref{fig:training_dynamics} denotes the inference accuracy of standard long-context reasoning when the maximum number of newly generated tokens is set to 8{,}192, which also corresponds to the maximum token length used during training for our method. 
As the number of iteration epochs increases, the accuracy of the proposed method gradually surpasses the baseline across benchmarks. This result indicates that the proposed method is able to overcome the context length limitation imposed during training and achieve higher inference accuracy beyond the fixed training window. 
At the same time, efficient training is maintained, as evidenced by the reduced training cost reported in Table~\ref{tab:training_time}.

\subsection{Ablation Study}

\begin{table*}[t]
\centering
\small
\setlength{\tabcolsep}{10pt}
\begin{tabular}{l cc cc cc cc}
\toprule
\multirow{2}{*}{\textbf{Segment Length}} 
& \multicolumn{2}{c}{\textbf{GSM8K}} 
& \multicolumn{2}{c}{\textbf{MATH500}} 
& \multicolumn{2}{c}{\textbf{AIME25}} 
& \multicolumn{2}{c}{\textbf{AMC23}} \\
\cmidrule(lr){2-3} \cmidrule(lr){4-5} \cmidrule(lr){6-7} \cmidrule(lr){8-9}
& ACC $\uparrow$ & LAT $\downarrow$
& ACC $\uparrow$ & LAT $\downarrow$
& ACC $\uparrow$ & LAT $\downarrow$
& ACC $\uparrow$ & LAT $\downarrow$ \\
\midrule
2K 
& \textbf{93.9} & 0.39 
& 82.4 & \textbf{1.95} 
& 20.6 & \textbf{5.87} 
& 59.7 & \textbf{4.03} \\

\rowcolor{Highlight}
4K 
& 93.6 & \textbf{0.34} 
& \textbf{86.0} & 2.19 
& \textbf{26.2} & 6.81 
& 64.3 & 4.30 \\

6K 
& 93.7 & 0.92 
& 84.2 & 3.28 
& 23.5 & 8.06 
& \textbf{64.7} & 4.90 \\
\bottomrule
\end{tabular}
\vspace{1mm}
\caption{Effect of segment length on accuracy (ACC) and latency (LAT) across benchmarks.
Higher ACC and lower LAT indicate better performance. Best results for each metric are highlighted in bold.}
\label{tab:context_length_ablation}

\vspace{-6mm}
\end{table*}

\textbf{Segment Length.}
Table~\ref{tab:context_length_ablation} reports the performance across benchmarks when different segmentation lengths (2K, 4K, and 6K) are used during training data construction, where 4K serves as the default setting. 
Across all four benchmarks, a segmentation length of 4K achieves the best or highly competitive accuracy. In addition, on MATH500, AIME25, and AMC23, the latency (LAT) increases as the segmentation length grows. 
This behavior is expected, since larger segmentation lengths gradually approach standard long-context reasoning, which incurs higher inference cost due to longer effective reasoning sequences.

\begin{table}[t]
\centering
\small
\setlength{\tabcolsep}{6pt}
\begin{tabular}{lccc}

\toprule
 & \textbf{AIME25} & \textbf{AMC23} & \textbf{GPQA-D} \\
\midrule
Baseline & 26.25 & 64.38 & 46.0 \\
w/o Image 
& 23.33\down{-11.1\%}
& 59.53\down{-7.5\%}
& 34.3\down{-25.4\%} \\
\bottomrule
\end{tabular}
\vspace{1mm}
\caption{Performance comparison with and without image input.
- denotes the relative performance drop.}
\label{tab:ablation_image_input}
\vspace{-4mm}
\end{table}

\textbf{Image Input.}
We further analyze the performance of \methodname when image inputs are removed at each reasoning iteration. 
Three more challenging benchmarks AIME25, AMC23, and GPQA-D, are selected for this analysis, which are more likely to benefit from multi-step reasoning.

As shown in Table~\ref{tab:ablation_image_input}, removing image inputs leads to accuracy drops of 11.1\% and 7.5\% on AIME25 and AMC23, with a more substantial degradation of 25.4\% observed on GPQA-D.
These results indicate that \methodname relies on rendered images as a form of memory for previous reasoning steps during inference. At the same time, a non-trivial level of accuracy is retained even without image inputs. This can be attributed to the fact that many problems can be solved within a single reasoning iteration; in the absence of image conditioning, the model effectively restarts the reasoning process from scratch and can still obtain correct answers.

% \textbf{attention map}

\begin{table}[t]
\centering
\resizebox{1.0\linewidth}{!}{
\begin{tabular}{lcc}
\toprule
\textbf{Method} & \textbf{Training Data} & \textbf{Training Time (h)} \\
\midrule
Base SFT & 60K QA pairs & 38.93 \\
SFT & 60K QA pairs & 38.92 \\
\methodname & 106K QA pairs + 105K images & \textbf{18.93} \\
\bottomrule
\end{tabular}
}

\caption{Training time comparison across different methods.}
\label{tab:training_time}
\vspace{-7mm}
\end{table}

\subsection{Efficiency Analysis}

\textbf{Training Efficiency.} Table~\ref{tab:training_time} reports the training time of \methodname in comparison with Base SFT and SFT. 
All training times are measured using the \texttt{LlamaFactory} framework under same configuration. 
Although the proposed method adopts a multi-iteration training paradigm and therefore introduces more QA pairs as well as additional images, the overall training time is reduced to approximately 48\% of that required by the baseline methods. 
This result demonstrates the training efficiency of \methodname. 
And the final performance of \methodname is superior as shown in Table~\ref{tab:math_benchmark_main}. The reduction in training time is attributed to the standard long-context reasoning involves substantially longer reasoning sequences, where training cost increases rapidly as the reasoning length grows. 
In contrast, \methodname constrains the reasoning length within each iteration to a controlled range, which leads to improved training efficiency.

% \begin{table}[t]
%     \centering
%     \caption{Statistics of VTC Rendering, calculated based on 100 samples.}
%     \label{tab:vtc_overhead}
%     \begin{tabular}{lc}
%         \toprule
%         \textbf{Metric} & \textbf{Value} \\
%         \midrule
%         Avg. Text Tokens per Image & 1.6k \\
%         Avg. Rendering Time & 0.12s \\
%         Avg. Image Processor Time & 0.02s \\
%         Avg. Image Size & 147KB \\
%         \bottomrule
%     \end{tabular}
% \end{table}

\textbf{Rendering Efficiency.}
Table~\ref{tab:math_benchmark_main} shows that \methodname significantly outperforms all baselines in terms of end-to-end latency, where the reported metric already accounts for the overhead of rendering and image processing. 
We further provide fine-grained statistics to validate that the introduced vision-text compression mechanism is lightweight. 
Based on an analysis of 100 samples from the dataset, we observe that for an average of approximately 1{,}600 text tokens per image, the rendering process requires only 0.12s on average, while image processing takes merely 0.02s. 
Compared to the overall model inference latency, this additional overhead is negligible (4\% of the total latency). 
Moreover, the average generated image size is around 0.1~MB, which falls within a practical and manageable range for real-world systems.

\subsection{Case Study}
We present four examples in Appendix~\ref{apd:case_study} to qualitatively analyze the behavior of \methodname. These examples illustrate that our method can condition on prior reasoning to perform solution verification, reasoning summarization, error correction based on identified contradictions, and direct continuation of preceding reasoning. Together, they demonstrate that images rendered from previous reasoning segments can be effectively leveraged to support multi-step reasoning.

\section{Conclusion}

We propose \methodname, an efficient long-context reasoning paradigm that integrates vision-text compression into iterative reasoning.
By rendering previous reasoning segments into compact visual representations, \methodname replaces long textual contexts with significantly fewer vision tokens in a lightweight and model-free manner.
Extensive experiments show that \methodname consistently improves reasoning accuracy across multiple benchmarks while achieving up to $3.4\times$ token compression and $2.7\times$ end-to-end inference speedup.
The results demonstrate that \methodname provides an effective alternative representation for scalable long-context reasoning. We hope our work would inspire further exploration of efficient reasoning beyond pure text-based paradigms.

% \section*{Acknowledgements}

\newpage
\section*{Impact Statement}

This paper presents work whose goal is to advance the field of LLMs Reasoning. There are many potential societal consequences of our work, none
which we feel must be specifically highlighted here.

\bibliography{example_paper}
\bibliographystyle{icml2026}

%%%%%%%%%%%%%%%%%%%%%%%%%%%%%%%%%%%%%%%%%%%%%%%%%%%%%%%%%%%%%%%%%%%%%%%%%%%%%%%
%%%%%%%%%%%%%%%%%%%%%%%%%%%%%%%%%%%%%%%%%%%%%%%%%%%%%%%%%%%%%%%%%%%%%%%%%%%%%%%
% APPENDIX
%%%%%%%%%%%%%%%%%%%%%%%%%%%%%%%%%%%%%%%%%%%%%%%%%%%%%%%%%%%%%%%%%%%%%%%%%%%%%%%
%%%%%%%%%%%%%%%%%%%%%%%%%%%%%%%%%%%%%%%%%%%%%%%%%%%%%%%%%%%%%%%%%%%%%%%%%%%%%%%
\newpage
\appendix
\onecolumn

\section{Image Rendering}

\subsection{Rendering Configuration}
\label{apd:rendering_config}

\begin{table}[h]
\centering
\small
\setlength{\tabcolsep}{6pt}
\begin{tabular}{p{0.24\textwidth} p{0.68\textwidth}}
\toprule
\textbf{Factor} & \textbf{Specification / Sampling Strategy} \\
\midrule
\texttt{dpi} & Mixture of sets: \emph{lowest} (45--59), \emph{low} (60--71), \emph{medium} (72--119), \emph{normal} (\{72, 80, 96, 100, 110, 120, 144, 150, 300\}), \emph{high} (over 300); favor normal/medium with small probability spikes to extremes. \\

\texttt{page\_size} & (i) Fixed paper sizes (A4, Letter, Legal, A5, B5, A3, B4, Tabloid) with priors; (ii) common aspect ratios (e.g., 1.414, 1.333, 1.5, 1.778); (iii) fully random aspect via piecewise distribution (narrow $\rightarrow$ tall). \\

\texttt{font\_family} & Pooled and deduplicated families across serif/sans/mono/pixel; italics sampled by filename heuristics (suffixes, \texttt{italic}/\texttt{oblique}). \\

\texttt{font\_size} & $\{7, 7.5, 8, 9, 9.5, 10, 11, 12, 14\}$; \texttt{line\_height} tied as \texttt{font\_size} + $\{0,\dots,3\}$. \\

\texttt{alignment} & \textsc{Left}/\textsc{Justify} (dominant) with small-prob.\ \textsc{Right}/\textsc{Center}. \\

\texttt{margins} & Three patterns: all-equal; vertical-larger; horizontal-larger; values in 10--40\,pt ranges. \\

\texttt{indent} & Modes: none; first-line indent ($\approx$1--2.5\,em); block/hanging with left/right indents. \\

\texttt{spacing} & \texttt{space-before}/\texttt{space-after} use a multi-mode prior (none, small, large). \\

\texttt{h\_scale} & Horizontal glyph scaling (0.75--1.0) with decaying probabilities. \\

\texttt{colors} & Page/background/font palettes for light/dark themes; document/web/code styles inherit coherent triplets (page, paragraph, font). \\

\texttt{borders} & Optional paragraph borders with width/padding; disabled by default. \\

\texttt{newline\_markup} & With small probability, explicit markers (e.g., \verb|\n|, tags, or tokens) inserted to preserve structure. \\

\texttt{auto\_crop} & Optional white-margin cropping and last-page trimming. \\
\bottomrule
\end{tabular}
\vspace{1mm}
\caption{Rendering configuration factors in the rendering pipeline and their sampling strategies.}
\label{tab:rendering-params}
\end{table}

The rendering pipeline is parameterized by a configuration vector. 
Following~\cite{glyph}, a set of rendering configuration factors is adopted, as summarized in Table~\ref{tab:rendering-params}. 
These factors determine the final rendering properties, including layout, visual clarity, and typography.

\begin{figure}[ht]
\centering
\begin{minipage}{0.48\linewidth}
    \centering
    
\begin{tabular}{p{0.45\linewidth} p{0.45\linewidth}}
\toprule
\textbf{Parameter} & \textbf{Value} \\
\midrule
\texttt{page-size} & 595 $\times$ 842 \\
\texttt{dpi} & 72 \\
\texttt{margin-x} & 10 \\
\texttt{margin-y} & 10 \\
\texttt{font-path} & \texttt{DejaVuSans.ttf} \\
\texttt{font-size} & 9 \\
\texttt{line-height} & 10 \\
\texttt{font-color} & \#000000 \\
\texttt{alignment} & LEFT \\
\texttt{horizontal-scale} & 1.0 \\
\texttt{first-line-indent} & 0 \\
\texttt{left-indent} & 0 \\
\texttt{right-indent} & 0 \\
\texttt{space-before} & 0 \\
\texttt{space-after} & 0 \\
\texttt{border-width} & 0 \\
\texttt{border-padding} & 0 \\
\texttt{page-bg-color} & \#FFFFFF \\
\texttt{para-bg-color} & \#FFFFFF \\
\texttt{auto-crop-width} & true \\
\texttt{auto-crop-last-page} & true \\
\bottomrule
\end{tabular}
\vspace{1mm}
\caption{Default rendering configuration used in our experiments.}
\label{tab:rendering-default}

\end{minipage}\hfill
\begin{minipage}{0.48\linewidth}
    \centering
    \includegraphics[width=\linewidth]{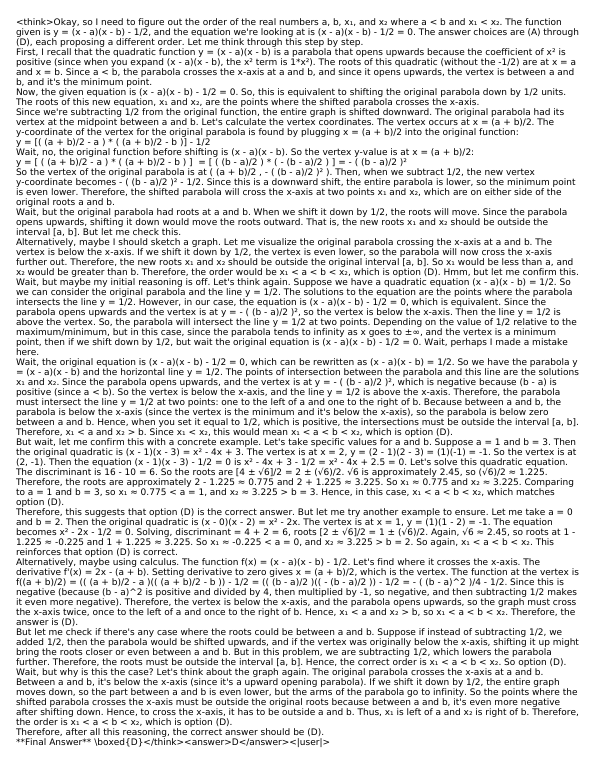}
    \captionof{figure}{Example rendered page.}
    \label{fig:example_rendering}
\end{minipage}

\end{figure}

The default configuration used in our experiments is reported in Figure~\ref{tab:rendering-default}. 
This configuration largely follows the default settings of Glyph. 
However, since the default Glyph font may produce incorrect glyphs when rendering certain mathematical symbols, the font is replaced with \texttt{DejaVuSans.ttf} in our implementation.

\subsection{Rendering Example}

Figure~\ref{fig:example_rendering} presents an example image rendered under the default configuration specified in Figure~\ref{tab:rendering-default}.

\section{Details about Experiments}

\subsection{Implementation Details.}
\label{apd:details}

Supervised fine-tuning is conducted using the \texttt{LlamaFactory} library~\cite{zheng2024llamafactory}. 
For all methods and across all model architectures, a learning rate of $1\times10^{-5}$ is used, with a warmup ratio of $0.1$ and a cosine learning rate schedule. 
Training is performed for one epoch with a batch size of $64$, the maximum sequence length is increased to $32{,}768$ tokens. All models are trained using 8 NVIDIA H20 GPUs with 96\,GB of memory.

We adopt the official implementation of TokenSkip~\cite{xia2025tokenskipcontrollablechainofthoughtcompression}, which supports compression ratios ranging from 0.6 to 0.9. We observe that training becomes unstable and collapses when the ratio is set to 0.6; therefore, we use a compression ratio of 0.8 in our experiments.

All evaluation experiments are conducted on a single NVIDIA H20 GPU with 96\,GB of memory. 
Inference is performed using the vLLM framework, with a temperature of $0.6$ and a top-$p$ value of $0.95$. 
For standard SFT, the maximum number of generated tokens (\texttt{max\_new\_tokens}) is set to $32{,}768$. 
For \methodname, the maximum number of generated tokens per iteration is set to $8{,}192$, and the maximum number of iterations is set to $8$.

\subsection{Benchmark}
\label{apd:subsec_bench}

\textbf{GSM8K}: A widely used benchmark for multi-step reasoning, consisting of 8,500 grade school math word problems, with a canonical test set of 1,319 problems.

\textbf{MATH500}: A challenging math dataset comprising 500 problems from high school math competitions.

\textbf{AIME25}: A benchmark dataset consisting of 30 challenging mathematical problems from the 2025 American Invitational Mathematics Examination.

\textbf{AMC23}: A challenging evaluation set comprising 50 problems from the 2023 American Mathematics Competitions, serving as a benchmark for competition-level mathematical reasoning.

\textbf{GPQA-Diamond}: A high-quality subset of the GPQA benchmark, with 198 complex graduate-level multiple-choice questions across various scientific domains. It serves as the out-of-distribution benchmark in our evaluation.

\subsection{Training Dataset}

We use OpenR1-Math-Inf~\cite{yan2025inftythink}, which is a subset of OpenR1-Math-220K~\cite{openr1}.
The OpenR1-Math-220k dataset, a large-scale benchmark for mathematical reasoning. It consists of 220k math problems, each accompanied by two to four reasoning traces generated by DeepSeek R1 for problems sourced from NuminaMath 1.5. All traces have been verified using Math Verify.

We first perform data cleaning on the OpenR1-Math-Inf dataset, resulting in 60,688 valid instances. 
In OpenR1-Math-Inf, for each instance, the original reasoning trace is partitioned into multiple segments based on a hyperparameter $\eta$, which controls the maximum token length of each segment. 
Following the data construction procedure of \methodname, this process yields a total of 106K training instances and approximately 105K rendered images.

For the final answer $A$, the special token sequence \texttt{<answer>} $A$ \texttt{</answer>} is used to facilitate answer extraction and to explicitly indicate the termination of the reasoning process. 
For instances consisting of more than one reasoning step, when $\texttt{step} > 1$, the intermediate supervision is formatted as
\texttt{<think>Got it, let's continue. \{step\_text\}</think>}.

\subsection{Batch Inference}
\label{apd:batch_inference}

\begin{algorithm}[h]
\caption{\methodname Batch Inference}
\label{algo:vtc_r1_batch}
\begin{algorithmic}
\STATE {\bfseries Input:} batch of questions $\mathcal{Q}=\{Q_1, \dots, Q_B\}$; initial images $\{\mathcal{I}^{\text{init}}_1, \dots, \mathcal{I}^{\text{init}}_B\}$; vision language model $M$; system prompt $\langle \texttt{S}_{v} \rangle$; rendering operator $R_\theta$; maximum iteration $T$
\STATE {\bfseries Initialize:} 
\STATE \quad Active request set $\mathcal{S} \leftarrow \{1, \dots, B\}$
\STATE \quad Current image sets $\mathcal{I}_k \leftarrow \mathcal{I}^{\text{init}}_k$ for all $k \in \{1, \dots, B\}$
\STATE \quad Final answers $\mathcal{A} \leftarrow \{ \emptyset \}_{k=1}^B$

\FOR{$t = 1$ to $T$}
    \IF{$\mathcal{S} = \emptyset$}
        \STATE \textbf{break}
    \ENDIF
    
    \STATE \textcolor{gray}{Batch Generation via vLLM:}
    \STATE Construct batch prompts $\mathcal{P} \leftarrow \{ (\langle \texttt{S}_{v} \rangle, Q_k, \mathcal{I}_k) \mid k \in \mathcal{S} \}$
    \STATE Obtain batch outputs: $\{O_k\}_{k \in \mathcal{S}} \leftarrow M(\mathcal{P})$
    
    \STATE \textcolor{gray}{Update States and Render:}
    \FOR{\textbf{each} $k \in \mathcal{S}$}
        \IF{$O_k$ produces the final answer}
            \STATE $A_k \leftarrow \text{ExtractAnswer}(O_k)$
            \STATE $\mathcal{S} \leftarrow \mathcal{S} \setminus \{k\}$ \COMMENT{Remove finished request from active set}
        \ELSE
            \STATE Extract reasoning progress $LR_k$ from $O_k$
            \STATE Render reasoning into images: $I_{\text{new}} \leftarrow R_\theta(LR_k)$
            \STATE Update image history: $\mathcal{I}_k \leftarrow \mathcal{I}_k \cup \{I_{\text{new}}\}$
        \ENDIF
    \ENDFOR
\ENDFOR

\IF{$\mathcal{S} \neq \emptyset$}
    \STATE \textcolor{gray}{Handle Time-out Requests:}
    \FOR{\textbf{each} $k \in \mathcal{S}$}
        \STATE $A_k \leftarrow \text{ExtractAnswer}(O_k)$
    \ENDFOR
\ENDIF

\STATE {\bfseries Output:} set of final answers $\mathcal{A}=\{A_1, \dots, A_B\}$
\end{algorithmic}
\end{algorithm}

\subsection{Case Study}
\label{apd:case_study}

The gray-shaded regions indicate reasoning steps that are performed by conditioning on images rendered from previous reasoning segments. Examples~1–4 are provided below. \textbf{Example~1} demonstrates further verification of a previously obtained solution. 
\textbf{Example~2} derives the final answer by summarizing completed prior reasoning. 
\textbf{Example~3} performs error correction and reflection based on contradictions identified in earlier reasoning, eventually reaching the correct answer. 
\textbf{Example~4} continues the reasoning process by building directly upon preceding reasoning steps. 
Collectively, these examples demonstrate that our method can successfully leverage images as \emph{optical memory} to support reasoning.

\begin{figure}[h]
\centering
\includegraphics[width=\textwidth]{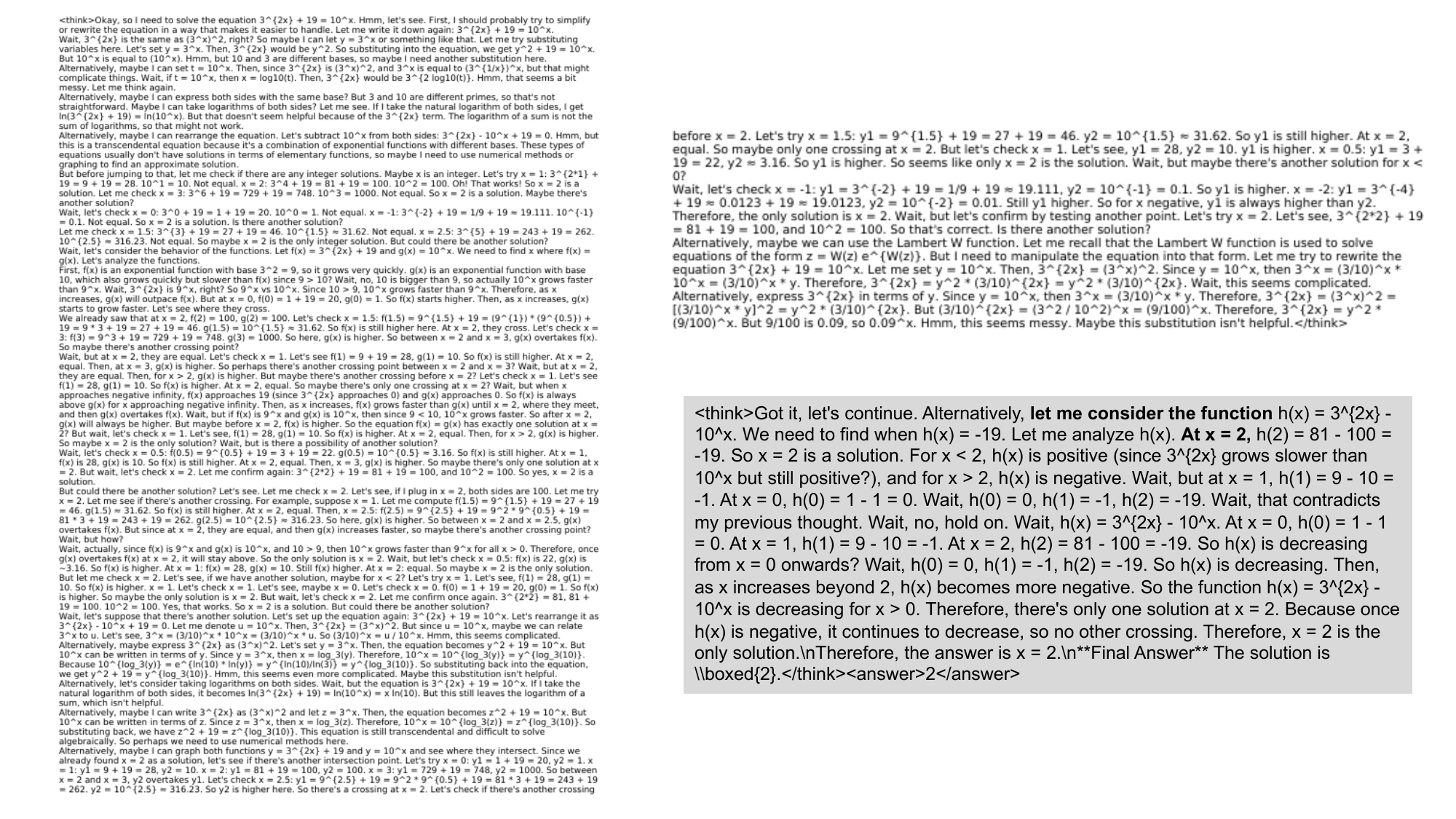}
\caption{Example 1.}
\label{fig:apd_case1}
\end{figure}

\begin{figure}[h]
\centering
\includegraphics[width=\textwidth]{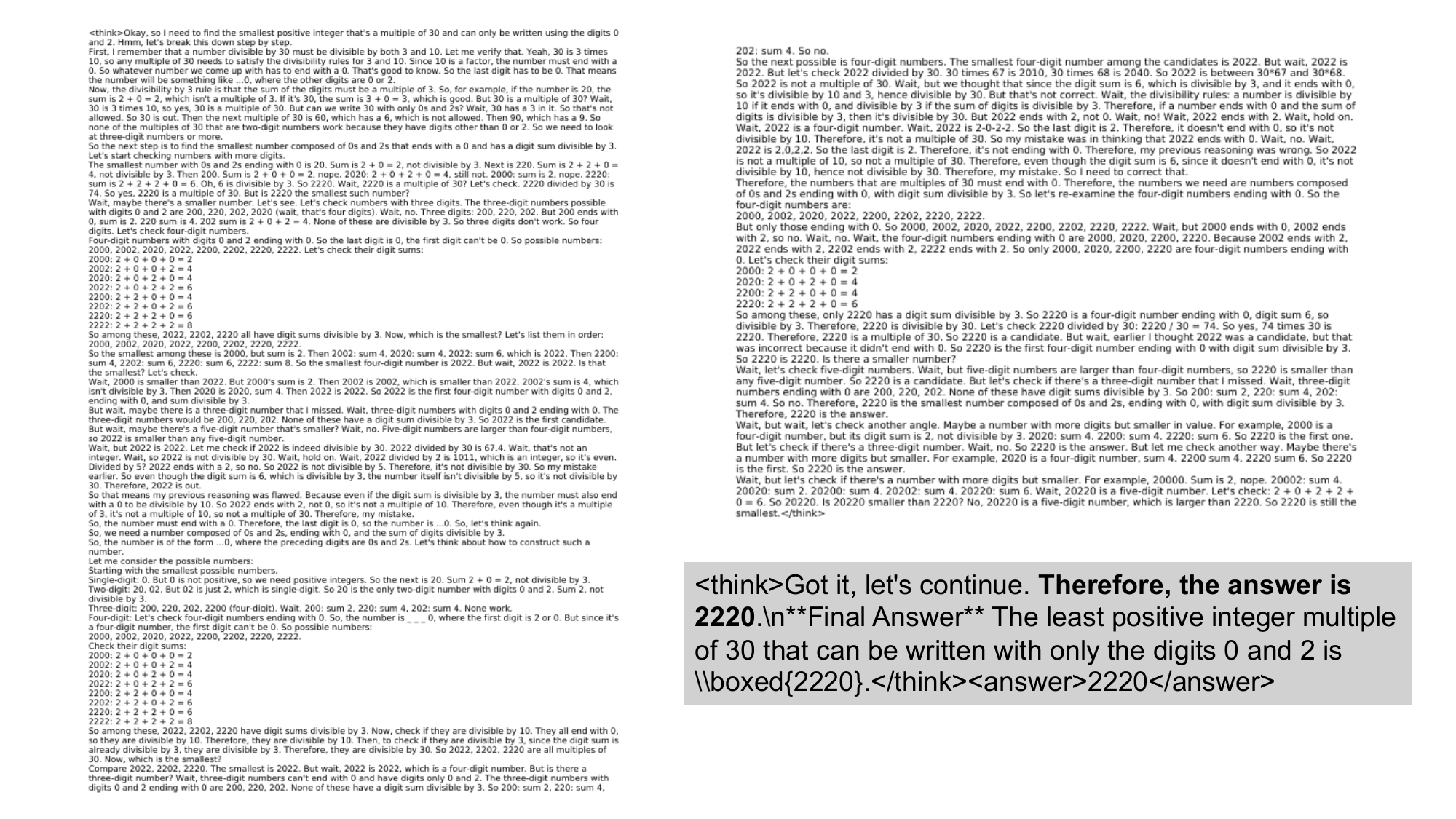}
\caption{Example 2.}
\label{fig:apd_case2}
\end{figure}

\begin{figure}[h]
\centering
\includegraphics[width=\textwidth]{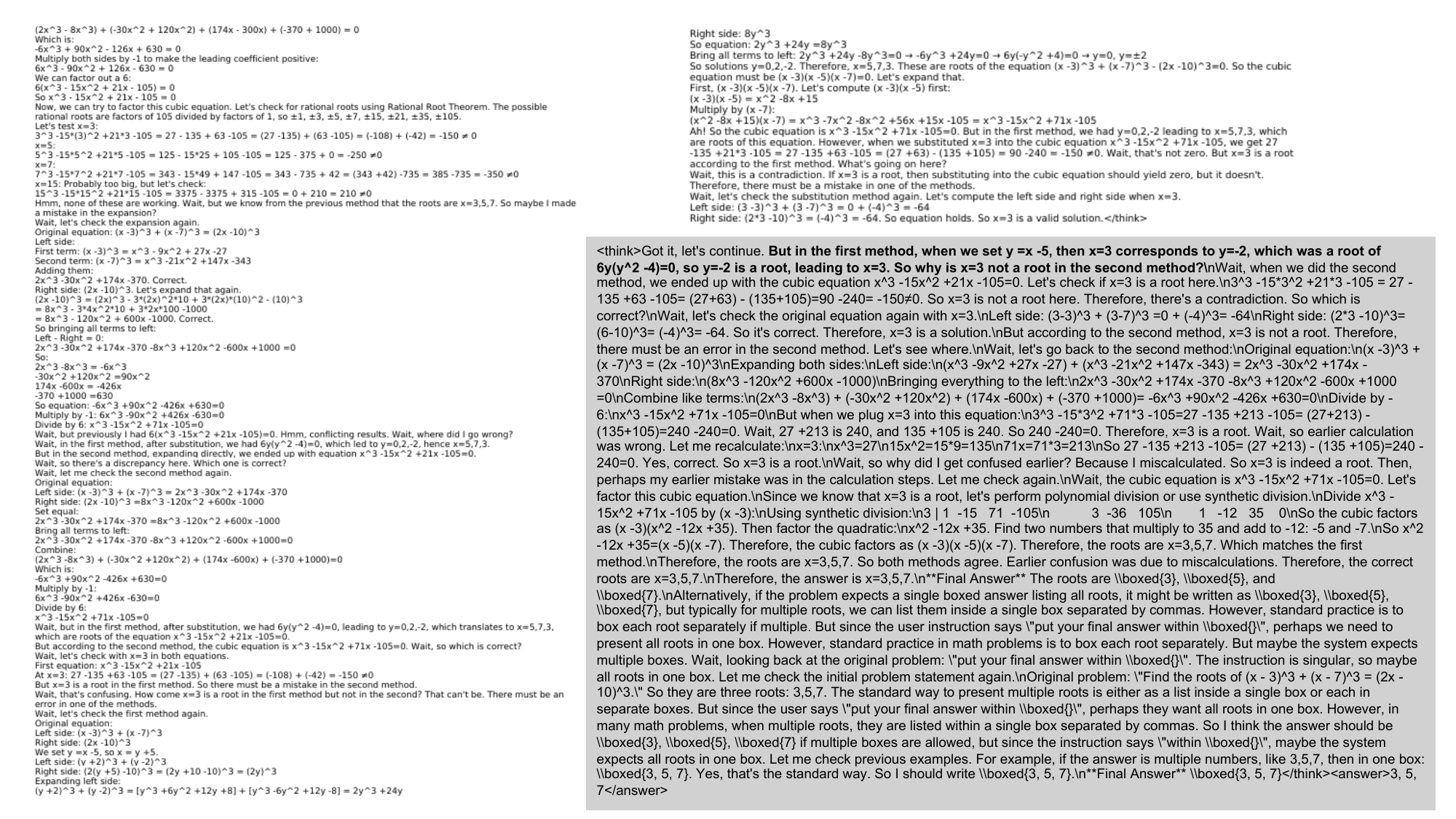}
\caption{Example 3.}
\label{fig:apd_case3}
\end{figure}

\begin{figure}[h]
\centering
\includegraphics[width=\textwidth]{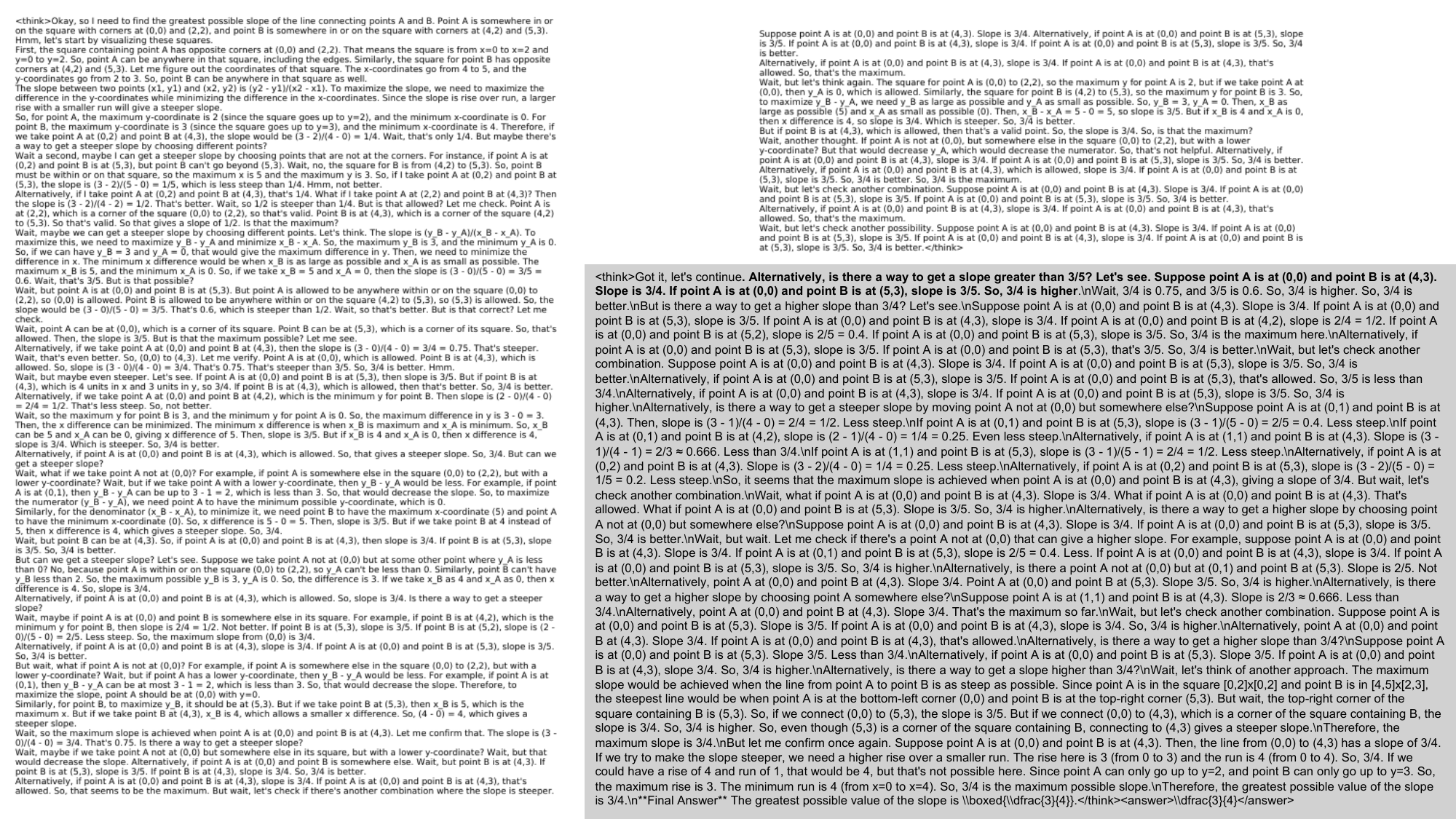}
\caption{Example 4.}
\label{fig:apd_case4}
\end{figure}

%%%%%%%%%%%%%%%%%%%%%%%%%%%%%%%%%%%%%%%%%%%%%%%%%%%%%%%%%%%%%%%%%%%%%%%%%%%%%%%
%%%%%%%%%%%%%%%%%%%%%%%%%%%%%%%%%%%%%%%%%%%%%%%%%%%%%%%%%%%%%%%%%%%%%%%%%%%%%%%

\end{document}